\pdfoutput=1

\documentclass[11pt]{article}

\usepackage[final]{acl}

\usepackage{times}
\usepackage{latexsym}
\usepackage{amsmath}
\usepackage{enumitem}
\usepackage{makecell}
\usepackage{multirow}
\usepackage{amsfonts}
\usepackage{booktabs}
\usepackage{cleveref}
\usepackage{tcolorbox}
\usepackage{float}
\usepackage[T1]{fontenc}
\usepackage[table]{xcolor}  
\definecolor{myblue}{RGB}{230, 242, 255}  
\definecolor{tableheadcolor}{gray}{0.92} 
\definecolor{tablerowcolor}{gray}{0.96} 

\usepackage[utf8]{inputenc}
\usepackage{CJKutf8}
\usepackage{microtype}
\usepackage{graphicx}

\usepackage{inconsolata}
\usepackage{graphicx}
\usepackage{subcaption}

\title{Do Large Language Models Perform Well on Comprehending Poetic Logic in Modern Chinese Poetry?}

\author{
Tian Lan$^{1}$\thanks{Equal contribution}, 
Shanshan Wang$^{2}$\footnotemark[1], 
Zehua Duo$^{3}$, 
Jiang Li$^{3}$, \\
\textbf{Guanglai Gao}$^{3}$,
\textbf{Derek F. Wong}$^{2}$, 
\textbf{Xiangdong Su}$^{3}$\thanks{\ \ Corresponding Author}
 \\
$^1$ Graduate School of Informatics, Kyoto University \\ 
$^2$ NLP²CT Lab, Department of Computer and Information Science, University of Macau\\
$^3$ College of Computer Science, Inner Mongolia University\\ 
\texttt{velikayascarlet@gmail.com, cssxd@imu.edu.cn}}

\usepackage{fontawesome5}

\begin{document}
\maketitle
\begin{abstract}

Large Language Models (LLMs) have achieved significant progress across a wide range of natural language processing (NLP) tasks, yet their ability to understand literary texts, particularly modern Chinese poetry, remains largely unexplored. The unique literary characteristics of modern Chinese poetry necessitate a distinct form of reasoning for effective comprehension. Unlike conventional texts that convey clear information, the unique "poetic logic" of modern Chinese poetry requires a holistic reasoning approach that goes beyond superficial semantic analysis to be understood. However, current evaluation paradigms largely ignore this critical dimension. To address this gap, we propose Peony, the first benchmark specifically designed for evaluating the poetic logic of modern Chinese poetry. We define poetic logic as four tasks across three levels, namely stanza, line, and imagery, and systematically evaluate and analyze six mainstream LLMs based on Peony. We evaluate these models under both non-thinking and thinking configurations. The experimental results reveal the limitations of current LLMs in understanding the poetic logic of modern Chinese poetry and validate the effectiveness and necessity of Peony. Our data and code will be available.

\end{abstract}

\section{Introduction}
\begin{figure*}[t]  
    \centering
    \includegraphics[width=0.99\textwidth]{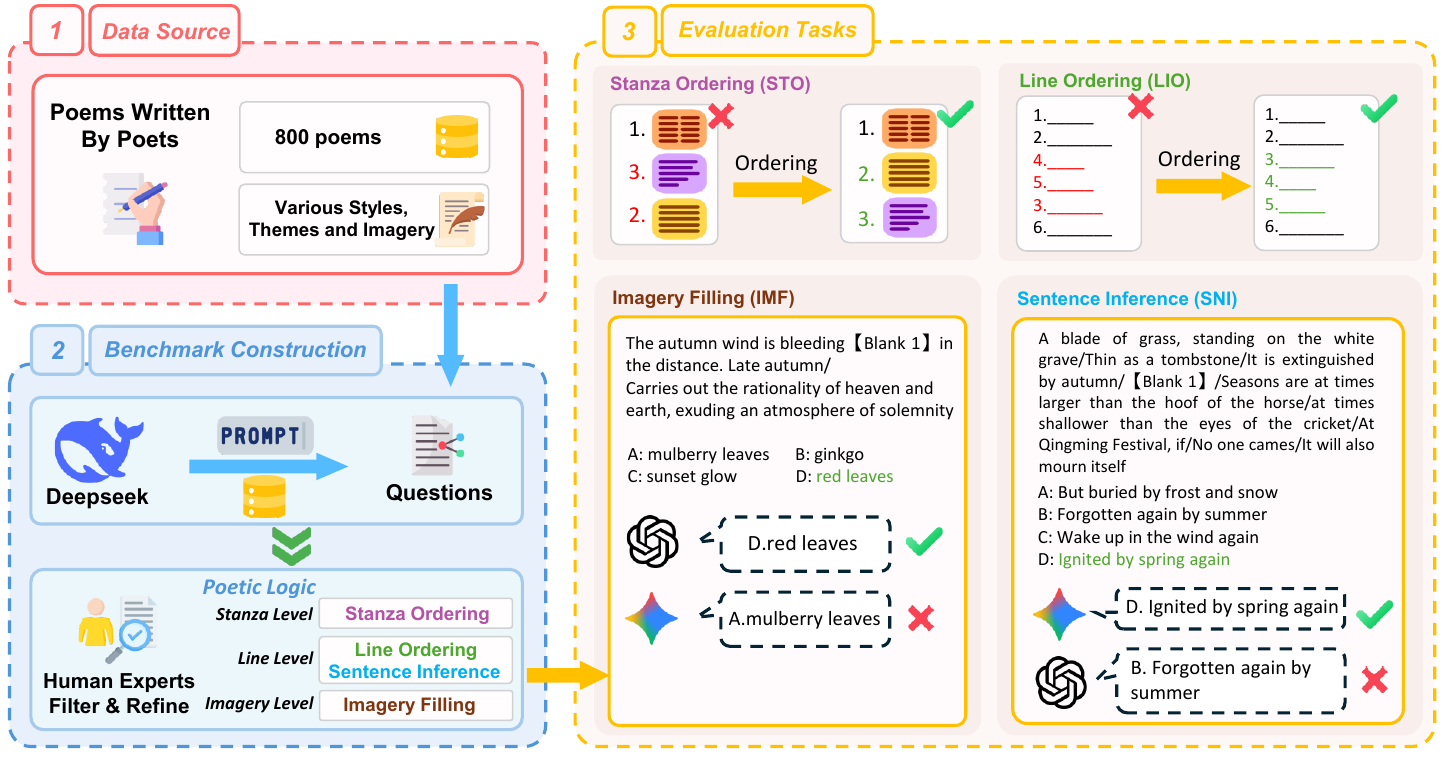} 
    \caption{The overall structure of our proposed benchmark.}
    \label{fig:overall} 
\end{figure*}

Large Language Models (LLMs) have achieved significant progress in numerous natural language processing (NLP) tasks, including reading comprehension~\citep{xiao2023evaluating}, summarization~\citep{liu2026calibrating,liu2026learning}, and code interpretation~\citep{peng2025swe}. While existing research has predominantly focused on general-domain language such as subcultural languages~\citep{lin2026exploring}, social media content~\citep{hu2025synergizing}, and programming languages~\citep{wang2026swe}, literary texts, particularly modern Chinese poetry, have received relatively little attention. The unique characteristics of modern Chinese poetry necessitate a distinct form of reasoning for effective comprehension.



Poetry represents one of the most complex forms of human language. Unlike conventional texts that primarily convey explicit information, poetry relies heavily on implicit associations~\citep{nahajec2009negation}, imagery~\citep{kao2012computational,wang2024research}, emotional resonance~\citep{johnson2022poetry}, and non-literal expressions~\citep{sinambela2025analyzing} to construct and communicate meaning. Consequently, comprehending poetry requires more than superficial semantic analysis; it demands the inference of implicit logical connections between lines, grounded in metaphorical associations~\citep{kovecses2013metaphor}, emotional progression~\citep{xia2021poetry}, and abstract semantics~\citep{wheelwright1940semantics}. Evaluating the mastery of this capability in LLMs provides a fresh perspective for evaluating whether these models truly possess profound language comprehension and abstract reasoning capabilities.

Among various poetry genres, modern Chinese poetry is particularly challenging. Unlike classical Chinese poetry, which typically adheres to fixed metrical structures~\citep{yu1971syntax,chen1979metrical}, modern Chinese poetry adopts freer forms and places greater emphasis on fragmented imagery, semantic ambiguity, emotional transitions, and unconventional syntax~\citep{skerratt2013form,wang2026can}, which constitutes a poetic logic that is distinct from the traditional texts.

\begin{CJK}{UTF8}{gbsn}



For instance, take the line "那等在季节里的容颜如莲花的开落" (the face waiting in the seasons is like the blooming and falling of a lotus) in the "\textit{《错误》(The Mistake)}" as an example, the poet does not explicitly state the character relationships or the temporal logic; instead, the poet implicitly conveys emotional shifts, such as waiting, the passage of time, and melancholy, by juxtaposing the images of "季节 (seasons)" and "莲花的开落 (the blooming and falling of a lotus)". Similarly, the expression "黑夜从大地上升起" (the night rises from the earth) in the poem "\textit{《黑夜的献诗》(A Poem Dedicated to the Night)}" breaks the conventional perception of the relationship between "夜(night)" and spatial dynamics in everyday language. These poetic characteristics, rooted in implicit imagistic associations and unconventional expressions, require readers to actively synthesize emotional and semantic connections to reconstruct the underlying poetic logic, which is the cue of a poem's holistic meaning. This reliance on global logical integration poses a barrier for LLMs, as they often struggle to move beyond local semantic features to grasp the cohesive structure of modern Chinese poetry.

However, this critical "poetic logic" remains largely ignored in current NLP evaluation paradigms. There is currently no specialized framework to evaluate whether LLMs truly comprehend the underlying logical structures of modern poetry. This omission prevents us from answering a fundamental question: do current LLMs truly possess the capability to comprehend the poetic logic of modern Chinese poetry?


To address this gap, we propose \textbf{Peony}, a \textbf{P}o\textbf{e}tic logic reaso\textbf{n}ing benchmark for modern Chinese poetr\textbf{y}. Peony aims to systematically evaluate the capability of current LLMs to comprehend the poetic logic of modern Chinese poetry. This benchmark contains 800 high-quality works by six contemporary poets, covering various themes, lengths, and writing styles to ensure thematic breadth and stylistic diversity. Furthermore, we formalize poetic logic into four tasks across three levels: line, stanza, and imagery. This allows us to systematically evaluate the capabilities of LLMs in the context of modern Chinese poetry comprehension. Based on this benchmark, we conduct comprehensive experiments and analyses on six mainstream models across two groups. The overall structure can be found in Figure \ref{fig:overall}.

\begin{figure*}[t]  
    \centering
    \includegraphics[width=0.99\textwidth]{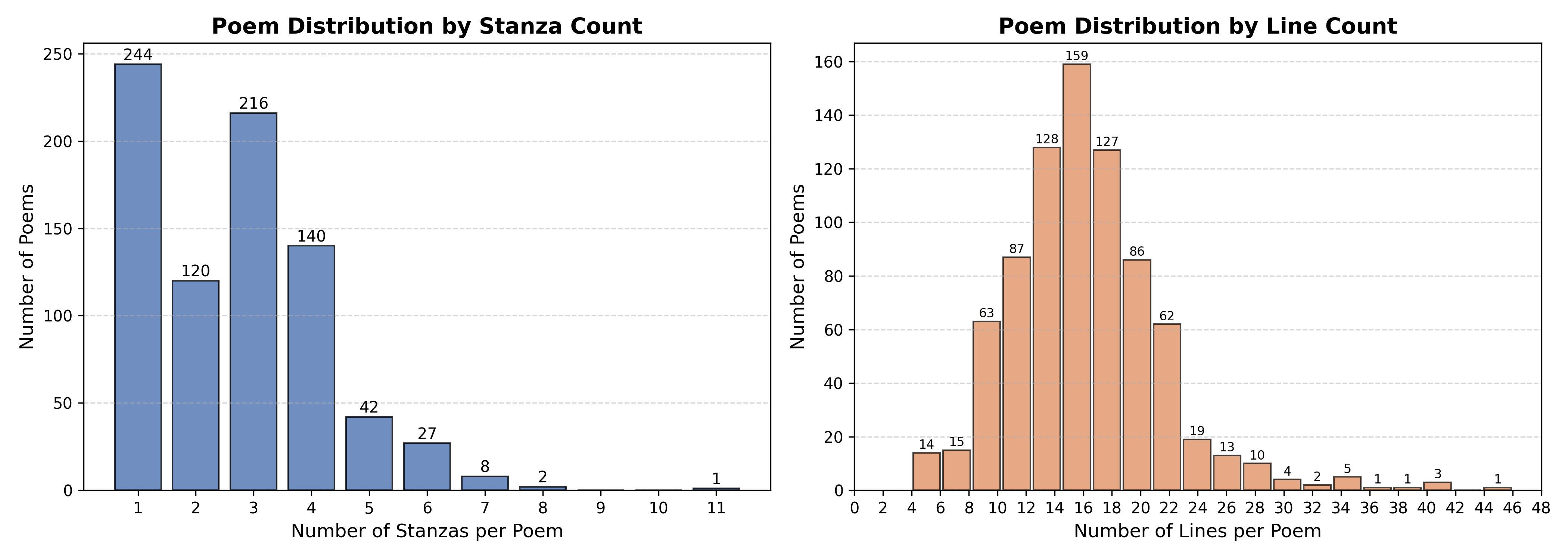} 
    \caption{The distribution of the poems categorized by the number of stanzas and lines.}
    \label{fig:stats} 
\end{figure*}

Our key contributions are as follows:

\end{CJK}

\begin{itemize}
    \item \textbf{Evaluation Benchmark} We introduce \textbf{Peony}, the first benchmark specifically designed for the poetic logic of modern Chinese poetry. It contains high-quality works by contemporary poets and covers various themes and styles.
    \item \textbf{Comprehensive Tasks} We have established the first evaluation benchmark tailored to the poetic logic of modern Chinese poetry, including four tasks across lines, stanzas, and imagery to comprehensively evaluate the performance of LLMs in comprehending the poetic logic of modern Chinese poetry.
    \item \textbf{Experimental Analysis} We conduct a detailed evaluation of popular LLMs on the Peony benchmark and provide a thorough analysis of the results. Our work offers new insights into the research concerning the comprehending of literary works by LLMs.
\end{itemize}

\section{Related Works}

\subsection{LLMs in Poetic Domains}
Poetry has gained widespread attention due to its unique stylistic features, immense literary value, and significant challenges. Current research primarily focuses on three directions: poetry generation, translation, and detection.

\textbf{Poetry Generation.} Since the pre-training corpora of LLMs contain a vast amount of poetry in different languages, the models themselves have demonstrated a certain capability for poetry generation~\citep{antar2023effectiveness,huang2025poembert,koziev2025generation}. Progress has been made regarding sentiment and style control during the generation process~\citep{shao2021sentiment} as well as format constraint issues~\citep{yu2024charpoet,song2025mixsong}. Furthermore, \citet{deng2024can} proposed an evaluation framework to measure the capability of LLMs to create classical Chinese poetry, while \citep{li2026poemetric} proposed POEMetric to evaluate the capability of LLMs to generate English poetry from the three dimensions of execution capability, creativity, and overall quality.

\textbf{Poetry Translation.} \citet{song2023towards} proposed a zero-shot poetry style injection method, treating poetry translation as a standard machine translation problem. \citet{resende2024translator} explored the potential of LLMs to improve the translation quality of rhymed and non-rhymed poetry. \citet{wang-etal-2024-best} proposed an explanation-based poetry machine translation method to enhance the performance of ChatGPT in poetry translation tasks. 

\textbf{Poetry Generation Detection.} \citet{wang-etal-2025-benchmarking} constructed a detection benchmark for modern Chinese poetry and discovered that existing detectors struggle to effectively identify poetry generated by large language models, while \citet{li2026wrote} focused on classical Chinese poetry and proposed the ChangAn benchmark, verified the limitations of current decision-based and probability-based detectors in this task.


\subsection{LLM's Comprehension of Poetry}

Poetic comprehension is a crucial prerequisite for advancing downstream tasks~\citep{yanfang2014translating,wang2026can} such as poetry generation, translation, and detection. \citet{zhao2024understanding} studied ancient Chinese poetry by aggregating multi-source corpora to design benchmarks that analyze literary nuances and evaluate model comprehension across various collections. \citet{jannidis2025large} conducted case studies and probing experiments to analyze how LLMs process German poetry across multiple linguistic dimensions, including meter, rhythm, vocabulary, syntax, and figurative language. \citet{al2025fann} introduced the Fann Or Flop benchmark to evaluate Arabic poetry comprehension across diverse genres and historical periods, evaluating capabilities in semantic interpretation, metaphor analysis, rhythmic awareness, and cultural context. Additionally, \citet{wang2026can} evaluated ChatGPT's capability to interpret modern Chinese poetry across multiple dimensions. Despite these efforts, the poetic logic underlying modern Chinese poetry remains underexplored, necessitating a more rigorous approach to evaluating model comprehension of modern Chinese poetry.


\section{The Dataset}

\subsection{Dataset Statistics}
There remains a lack of systematic evaluation benchmarks for the comprehension of poetic logic in modern Chinese poetry. To address this gap, we construct the Peony Benchmark based on existing modern Chinese poetry corpora, aiming to evaluate the capabilities of LLMs in comprehending the poetic logic of modern Chinese poetry. Specifically, we utilize the high-quality modern Chinese poetry created by humans from the AIGenPoetry dataset~\citep{wang-etal-2025-benchmarking} as a foundation, and further design and construct tasks, questions, and ground-truth answers focused on poetic logic to form a new benchmark for evaluating the poetic logic of modern Chinese poetry. Peony contains 800 modern Chinese poems in total. The statistical information of the benchmark is presented in Table~\ref{tab:peony_stats}, and the distribution of the poems categorized by the number of stanzas and lines is illustrated in Figure~\ref{fig:stats}.

The Peony benchmark comprises 800 modern Chinese poems by six contemporary poets, curated from the AIGenPoetry dataset. Table~\ref{tab:peony_stats} reports the corpus and task statistics, while Figure~\ref{fig:stats} shows the distribution of poems by numbers of stanzas and lines.

The selection of contemporary works over canonical poetic works aims to mitigate data leakage risks. Given that most canonical poetic works are likely included in the pre-training corpora of LLMs, models may succeed through memorization or pattern retrieval, thereby obscuring their actual capability to comprehend poetic logic. AIGenPoetry was released in November 2025, after the reported knowledge cutoff dates of all evaluated models. 

\begin{table}[t]
\centering
\rowcolors{2}{white}{tablerowcolor} 
\begin{tabular}{lc}
\toprule
\rowcolor{tableheadcolor} 
\textbf{Metric} & \textbf{Value} \\
\midrule
Number of Poems & 800 \\
Number of lines & 12,826 \\
Average lines per Poem & 16.03 \\
Average Characters per line & 10.23 \\
Number of LIO & 800~(800) \\
Number of STO & 436~(436) \\
Number of IMF & 400~(695) \\
Number of SNI & 800~(2424) \\
\bottomrule
\end{tabular}
\caption{Statistical information of the Peony benchmark. The actual number of questions is indicated in parentheses.}
\label{tab:peony_stats}
\end{table}

\subsection{Benchmark Construction}

The construction of Peony follows a systematic pipeline. We first select task-compatible samples from the 800 poems and then apply line/stanza shuffling or imagery/sentence masking according to each task. Following prior work~\citep{huang2024cbbq,lan-etal-2025-f2bench}, we use DeepSeek-V4-Pro~\citep{deepseekai2026deepseekv4} to generate initial mask positions and distractor candidates for IMF and SNI. Professional poets subsequently review all candidates and substantially revise or replace the vast majority before inclusion in the benchmark.

The annotators are two professional poets and poetry-theory researchers. For STO and LIO, they redesigned options containing obvious contextual breaks or contradictory imagery so that distinctions primarily depend on internal poetic logic. For IMF, they removed cases in which multiple imagery options could reasonably fill a blank and ensured a unique intended answer. For SNI, experts redesigned distractors that reused the same imagery as the correct answer, so that selecting an option required evaluating the coherence of the entire sentence with the poem rather than matching individual imagery words. These procedures and the expert revision process are described further in Appendix~\ref{app:annotation-quality}. The complete curation prompts are provided in Appendix~\ref{app:prompts}.

\subsection{Diversity}
Peony exhibits diversity in poetic form, content, and style. In terms of form, it includes single-stanza and multi-stanza poems, poems with uniform or varying numbers of lines per stanza, and symmetrical structures in which the opening and closing stanzas contain the same number of lines. The poems range from fewer than ten lines to more than thirty lines. In terms of content, the collection covers emotions, daily life, social issues, travel, philosophy, art, and artificial intelligence. Stylistically, it includes both poems that retain rhythmic and structural echoes of classical Chinese poetry and free-verse works characterized by fragmented syntax and unconventional imagery. A comparison with existing Chinese poetry benchmarks is provided in Appendix~\ref{app:supplementary-analysis}.
The full list of high-frequency imagery words is provided in Appendix~\ref{app:imagery}.


\section{Evaluation Tasks} 
Evaluating the capability of LLMs to comprehend the poetic logic of modern Chinese poetry presents significant challenges. Therefore, we introduce four tasks from three levels (line, stanza and imagery) in Peony to systematically measure it: \textbf{Stanza Ordering}, \textbf{Line Ordering}, \textbf{Imagery Filling}, and \textbf{Sentence Inference}. Each task is scored based on accuracy, ranging from 0 to 100, where higher scores indicate superior comprehension capabilities. Illustrative examples for each task are provided in Appendix~\ref{app:examples-of-poetry}. All evaluation items and their ground-truth answers are reviewed and finalized by professional poets with expertise in poetic theory and practice.

Given the high stylistic freedom of modern Chinese poetry, a single set of verses may allow for multiple logical interpretations. To avoid the ambiguity inherent in these tasks, we have designed this task as a multiple-choice format. Consequently, the model must make a clear and definitive selection from the provided candidates. The complete evaluation prompts for all four tasks are provided in Appendix~\ref{app:prompts}.

\begin{CJK}{UTF8}{gbsn}

\subsection{Stanza Ordering~(STO)}

The Stanza Ordering task evaluates a model’s capability to reconstruct the original logical sequence of a poem from shuffled stanzas. A stanza is a unit consisting of several continuous lines, typically separated by blank lines within the poem. Unlike the Line Ordering task, each stanza functions as a relatively coherent semantic unit. The logical progression between stanzas, such as narrative advancement, emotional development, and imagery evolution, is inherently more stable than the relationships between individual lines. In this task, we evaluate model performance using accuracy.

\begin{table*}[t]
  \centering
  \small
  \resizebox{0.9\textwidth}{!}{ 
  \begin{tabular}{llccccc}
    \hline
    \rowcolor{blue!15} \textbf{Category} & \textbf{Models} & \textbf{STO} & \textbf{LIO} & \textbf{IMF} & \textbf{SNI} & \textbf{Average} \\
    \hline 
    \multirow{5}{*}{\rotatebox{0}{\textbf{Non-thinking}}} 
    & DeepSeek-V4-Flash   & 47.02 & 49.31 & 66.08 & 37.21 & 49.91 \\
    \rowcolor{blue!5} & DeepSeek-V4-Pro   & 45.96 & 51.00 & \underline{72.93} & 42.12 & 53.01 \\
    & GPT-5.1             & 39.08 & 43.25 & 69.44 & 36.14 & 46.98 \\
    \rowcolor{blue!5} & Kimi-K2.5         & \textbf{52.38} & \textbf{63.00} & \textbf{75.25} & \textbf{48.93} & \textbf{59.89} \\
    & GLM-5.1             & \underline{49.54} & \underline{56.12} & 68.98 & \underline{42.20} & \underline{54.21} \\
    \hline 
    \multirow{5}{*}{\rotatebox{0}{\textbf{Thinking}}} 
    & DeepSeek-V4-Flash   & 50.00 & 67.59 & 76.09 & 39.81 & 58.37 \\
    \rowcolor{blue!5} & DeepSeek-V4-Pro   & 54.59 & 76.60 & \textbf{79.88} & 45.94 & 64.25 \\
    & Gemini 3 Pro        & \textbf{68.35} & \textbf{85.29} & 78.85 & \textbf{63.53} & \textbf{74.01} \\
    \rowcolor{blue!5} & Kimi-K2.5         & 56.88 & 71.50 & \underline{79.83} & \underline{48.85} & \underline{64.27} \\
    & GLM-5.1             & \underline{57.64} & \underline{77.69} & 76.48 & 44.93 & 64.19 \\
    \hline 
  \end{tabular}}
  \caption{Overall performance comparison of LLMs on the Peony benchmark (accuracy). \textbf{Bold} indicates the best performance and \underline{underline} indicates the second best.}
  \label{tab:peony-overall-performance}
\end{table*}

\subsection{Line Ordering~(LIO)} 

The core of the Line Ordering task is to evaluate a model’s capability to discern the underlying logical sequence of a text. As noted by \citep{guo2025c3lrso}, existing sentence-ordering datasets often contain explicit cue words that make logical relationships between sentences apparent. This allows models to achieve high scores by simply identifying these markers rather than comprehending the underlying text logic. In contrast, modern Chinese poetry rarely features such explicit markers; it is a genre that relies heavily on imagery associations and spatiotemporal discontinuity. Therefore, leveraging these characteristics of modern Chinese poetry enables a more accurate evaluation of whether a model truly understands the poetic logic unique to this genre.

In this task, the model is presented with two components: a set of shuffled sentences and a piece of contextual poetic text. The contextual text provides a complete poetic framework, encompassing elements such as imagery, mood, and intentional ambiguity. The model must utilize this context to reorder the shuffled sentences so that they align with the poetic logic of the provided poem.


\subsection{Imagery Filling (IMF)}

Imagery is an essential component of modern Chinese poetry, as poets often convey complex emotions and philosophical reflections through the layering, juxtaposition, and resonance of imagery. Based on this, we design the Imagery Filling task to evaluate whether a model can identify the most contextually appropriate imagery from multiple candidates within a given poetic logistical framework.

Specifically, we select a set of complete modern Chinese poems and designate several slots to be masked, with each slot corresponding to a multiple-choice question focused on imagery. 

The final score is calculated as the accuracy across all questions. This is expressed as follows:
\begin{equation}
Score =\sum_{i=1}^{M} \frac{C_i}{N_i} \times 100,
\end{equation} 
where $C_i$ is the number of correctly answered questions for the $i$-th poem, $N_i$ is the total number of questions for that poem, and $M$ is the total number of poems in the evaluation set. 

Furthermore, we report the \textit{Perfect Rate (PR)}, which is represented as the proportion of poems in which all blanks are correctly filled. The \textit{PR} is defined as:
\begin{equation}
PR =  \sum_{i=1}^{M}\frac{P_i}{M} \times 100
\end{equation}
where $P_i=1$ represents all questions in a poem are answered correctly and $P_i=0$ otherwise . This metric evaluates the model's consistency in maintaining logical coherence throughout an entire poetic work, regardless of whether a poem contains a single question or multiple questions.

\subsection{Sentence Inference (SNI)} 

The Sentence Inference task is designed to evaluate a model’s capacity to grasp the poetic logic connecting individual lines within modern Chinese poetry. Unlike the Imagery Filling task, which focuses on localized imagery collocation, Sentence Inference prioritizes inter-line coherence, encompassing diverse poetic logical relationships such as causality, contrast, progression, resonance, and imagery leaps. Modern Chinese poetry frequently deviates from conventional line-level connectors, such as chronological or causal chains, and instead relies on emotional flow, imagery accumulation, or intentional gaps to advance the text. Consequently, understanding imagery at the word or phrase level is insufficient; the model must also comprehend the poetic logical trajectory between lines.

Specifically, we select a set of complete modern poems and designate several critical positions to be masked. Each masked position corresponds to a multiple-choice question where the model must select the most contextually appropriate sentence from a set of candidates. Only one candidate is the original sentence from the poem, while the remaining options serve as distractors that are inconsistent with the underlying poetic logic. The masked content can be a complete line, or just a part of a line. The scoring criteria for this task are identical to those of the Imagery Filling task.

\end{CJK}

\section{Experimental Setup}

\paragraph{Models} We evaluate the performance of the models under two distinct settings. \textbf{Non-thinking mode}, which includes DeepSeek-V4-Flash and Pro, GPT-5.1~\citep{singh2025openai}, Kimi-K2.5~\citep{team2025kimi}, and GLM-5.1~\citep{zeng2026glm}. \textbf{Thinking mode}, which includes Gemini 3 Pro~\citep{pichai2025new}, as well as the aforementioned models (DeepSeek-V4-Flash and Pro, Kimi-K2.5, and GLM-5.1) operating with their respective reasoning modes activated. We adopt developer-recommended default parameters for all models.

\paragraph{Metrics} For LIO and STO tasks, we report Accuracy, and for IMF and SNI tasks, we report both Accuracy and \textit{PR}. All experiments are conducted three times, and we report the average values as our final results in Tables \ref{tab:peony-overall-performance} and \ref{tab:peony-perfect-rates}.

\section{Results and Analysis}

Figure \ref{Experimental examples} presents the experimental examples for different tasks. 


\subsection{Overall Performance}
\begin{CJK}{UTF8}{gbsn}

Table \ref{tab:peony-overall-performance} shows that the performance of different LLMs varies significantly. Specifically, among the non-thinking models, Kimi-K2.5 achieves the best and most balanced performance. Its average accuracy (59.89) is 5.68 points higher than that of the second-best non-thinking model, GLM-5.1 (54.21). This indicates that Kimi-K2.5 possesses a strong capability to capture poetic logic without explicit reasoning, especially in the Imagery Filling (IMF) and Sentence Ordering (LIO) tasks. In contrast, the worst performing non-thinking model is GPT-5.1, with an average score of 46.98.

Table \ref{tab:peony-perfect-rates} reports the \textit{PR} for IMF and SNI tasks. A consistent pattern emerges across both thinking and non-thinking models: the \textit{PR} values are substantially lower than the corresponding accuracy scores. For instance, DeepSeek-V4-Pro (Non-thinking) achieves an IMF accuracy of 72.93 but a \textit{PR} of only 59.95, indicating that models can handle individual imagery items but fail to coordinate multiple imagery units within a single poem. This gap is even more pronounced in the SNI task, where the highest \textit{PR} among all models is only 30.77 (Gemini 3 Pro), confirming that cross-sentence poetic logic remains a major challenge for current LLMs.

We additionally evaluate five professional poets and four non-expert readers on a random sample of 40 poems. Experts achieve higher average performance than non-experts, although performance varies substantially across individual readers. Full results and participant definitions are reported in Appendix~\ref{app:human-performance}.


GPT-5.1 performs the worst on all tasks except IMF. Our qualitative analysis suggests that it often relies on surface meaning and has difficulty integrating implicit themes, emotions, and Chinese cultural references. In a representative example, \textit{《庄子与倒影》(Zhuangzi and Reflection)}, its response does not identify the connection between Zhuangzi's Butterfly Dream, ``庄子 (Zhuangzi),'' and ``秋水 (Autumn Waters),'' resulting in an incorrect interpretation of the poem's theme. More generally, our qualitative analyses identify three recurring sources of model errors: reliance on surface meaning rather than underlying themes, difficulty following emotional progression, and failure to integrate interacting images into a coherent artistic conception. The SNI case study in Appendix~\ref{app:sni-case-study} illustrates how these limitations can occur together.

Overall, thinking configurations generally outperform their non-thinking counterparts. Kimi-K2.5 is the only exception on SNI, where the difference is small (48.85 versus 48.93). Gemini 3 Pro achieves the strongest overall performance (74.01), exceeding the second-best thinking model, Kimi-K2.5 (64.27), by 9.74 points and the best non-thinking model, also Kimi-K2.5 (59.89), by 14.12 points.

On the two tasks for which DeepSeek-V4-Pro assisted initial candidate generation, the DeepSeek models do not consistently outperform the other systems. DeepSeek-V4-Pro obtains 45.94 on SNI, below Gemini 3 Pro (63.53) and within the range of the other models; on IMF, its 79.88 is close to Kimi-K2.5 (79.83) and Gemini 3 Pro (78.85). These results provide no clear evidence of a substantial performance advantage associated with the annotation source.

The largest average gains from thinking mode are observed for DeepSeek-V4-Flash and DeepSeek-V4-Pro, which improve by 18.46 and 11.24 points, respectively, over their non-thinking configurations.

Notably, Gemini 3 Pro achieves its best performance on the LIO task (85.29), followed by the IMF task. Conversely, all other models, regardless of whether they are thinking or non-thinking, perform best on the IMF task, followed by the LIO task.

\begin{table}[t]
  \centering
  \resizebox{\columnwidth}{!}{ 
  \begin{tabular}{llccc}
    \hline
    \rowcolor{blue!15} \textbf{Category} & \textbf{Models} & \textbf{IMF \textit{PR}} & \textbf{SNI \textit{PR}} \\
    \hline 
    \multirow{5}{*}{\rotatebox{0}{\textbf{Non-thinking}}} 
    & DeepSeek-V4-Flash   & 51.27 & 11.62 \\
    \rowcolor{blue!5} & DeepSeek-V4-Pro   & \underline{59.95} & 12.50 \\
    & GPT-5.1             & 56.60 & 8.88  \\
    \rowcolor{blue!5} & Kimi-K2.5         & \textbf{62.50} & \textbf{18.62} \\
    & GLM-5.1             & 54.89 & \underline{13.12} \\
    \hline 
    \multirow{5}{*}{\rotatebox{0}{\textbf{Thinking}}} 
    & DeepSeek-V4-Flash   & 62.56 & 9.38  \\
    \rowcolor{blue!5} & DeepSeek-V4-Pro   & \textbf{69.10} & 15.77 \\
    & Gemini 3 Pro        & 66.75 & \textbf{30.77} \\
    \rowcolor{blue!5} & Kimi-K2.5         & \underline{67.76} & \underline{17.71} \\
    & GLM-5.1             & 61.78 & 14.79 \\
    \hline 
  \end{tabular}}
  \caption{\textit{PR} for IMF and SNI tasks across different models. \textbf{Bold} indicates the best performance and \underline{underline} indicates the second best.}
  \label{tab:peony-perfect-rates}
\end{table}

\subsection{Specific Tasks}

Experimental results show that the most challenging task is SNI, followed by STO, while the relatively easiest task is IMF.

\paragraph{Stanza Ordering (STO)} STO requires the model to grasp the understanding of the entire poem, thereby recovering the global structure of the stanzas. Gemini 3 Pro performs relatively best, with an accuracy of 68.35. Among non-thinking models, Kimi-K2.5 performs best (52.38). Experimental results show that all models perform worse on STO than on LIO, indicating that while stanza-level logic is more stable than line-level logic, it requires a higher capability from the model to grasp complete structural units, themes, and emotions.

\paragraph{Line Ordering (LIO)} LIO evaluates the model's capability to recover the logical order of lines within a poem. Gemini 3 Pro leads by a wide margin with an accuracy of 85.29, followed by GLM-5.1-Thinking (77.69) and DS-Pro-Thinking (76.60). Among non-thinking models, Kimi-K2.5 ranks first with 63.00, but still far below thinking models. It is noteworthy that the overall performance of non-thinking models is low, indicating that sentence-level poetic ordering is a significant challenge for most models. Taking "\textit{《伤痕》 (The Scar)}" as an example, the correct order B (15423) requires the model to recognize the emotional progression from the memory of "外公 (grandfather)" triggered by "伤疤 (scar)" to "小窝召唤故人 (the small nest calling back old friends)," a logic that non-thinking models struggle to understand. Conversely, thinking models can better infer the logical progression between lines through reasoning.

\paragraph{Imagery Filling (IMF)} Among non-thinking models, Kimi-K2.5 performs best, with both accuracy (75.25) and \textit{PR} (62.50) significantly higher than DeepSeek-V4-Flash (66.08 \& 51.27). Among the thinking models, DeepSeek-V4-Pro achieves the highest accuracy and \textit{PR} (79.88), followed by Kimi-K2.5. Compared to non-thinking models, the accuracy differences among different thinking models on the IMF task are relatively small.

Furthermore, the gap between accuracy and \textit{PR} is prevalent across all models. For example, while DeepSeek-V4-Pro's accuracy rate is 79.88, its \textit{PR} is only 62.56, meaning that at least one imagery is incorrectly filled in 37.44\% of the poems. Taking "\textit{《晚秋帖》 (Late Autumn Poem)}" as an example, the correct answer (D: 红叶 (red leaves), A: 月光 (moonlight), B: 秋虫 (autumn insects)) requires the model to simultaneously grasp the connection between three imagery items and the somber atmosphere of autumn. Any error in any one of these imagery compromises the \textit{PR}. This demonstrates that even the best-performing model struggles to achieve a coherent grasp of the overall imagery of the entire poem. The model's capability to handle individual blanks is acceptable, but its performance deteriorates when it needs to process the logic between multiple imagery simultaneously. This also reveals a current lack of understanding of poetry as a whole in LLMs.

\paragraph{Sentence Inference (SNI)} SNI is the most challenging among the four tasks, requiring the model to select the most appropriate sentence to maintain the consistency of the poetic logic within the poem. Gemini 3 Pro shows a significant lead, with an accuracy of 63.53 and a \textit{PR} of 30.77, far exceeding other models. We observed that this is because Gemini 3 Pro can accurately understand the meaning of the entire poem and the core themes and emotions it wants to convey, especially by correctly inferring the inherent logical relationships between candidate lines and other lines. 

Furthermore, Gemini 3 Pro also pays attention to the context surrounding the candidate lines. Taking "\textit{《一棵草》 (A Blade of Grass)}" as an example, Gemini 3 Pro correctly captured the contextual consistency between "被春天点燃 (ignited by spring)" and the cycle of seasons, "清明 (Qingming)" and the ritual of offering sacrifices, and the logical causal relationship of "祭奠自己 (mourn itself)" because "无人到来 (no one comes)." Although Gemini 3 Pro performs this task relatively well, its \textit{PR} is only 30.77, indicating that at least one sentence in 69.23\% of the poems is incorrectly selected.

Among the non-thinking models, Kimi-K2.5 performs relatively best, with an accuracy of 48.93 and a \textit{PR} of 18.62. We observed that Kimi-K2.5 could identify the theme and atmosphere of the poem through external imagery, but could not understand the deeper meaning and true emotions of the poem.



In summary, our results and analysis reveal distinct differences among various LLMs in their capacity for poetic logic, providing a reference for future evaluations of LLMs' performance in the domain of literary understanding.
\end{CJK}

\section{Conclusion}
In this paper, we propose Peony, the first benchmark specifically designed for evaluating the poetic logic of modern Chinese poetry. Peony comprises 800 modern Chinese poems, meticulously crafted by modern poets, featuring various poetic imagery, styles and themes. 
We systematically evaluated many popular LLMs with Peony, and our results show that all evaluated models exhibit substantial limitations in recovering poetic logic across the four Peony tasks. 
While all thinking models outperform their counterparts in average, the performance advantage of thinking models in the SNI task is not pronounced, suggesting that explicit reasoning strategies are not always applicable to the comprehension of poetic logic. 
Furthermore, although all models achieve relatively high accuracy on the IMF task, their \textit{PR} values are significantly lower, reflecting a pattern where models can manage individual imagery but fail to coordinate the logical relationships between multiple imagery units. 
By introducing Peony into the NLP community, we have opened up new research avenues in the evolving field of modern Chinese poetry comprehension.

\section*{Limitations}
Peony has several limitations. First, it contains 800 works by six contemporary poets. Although the poets span different styles, themes, and poetic forms, this collection cannot represent the full historical and stylistic diversity of modern Chinese poetry. Future versions should incorporate more poets and poems. Second, Peony operationalizes poetic logic through four tasks centered on structural organization, imagery association, emotional progression, and contextual coherence. It does not directly evaluate the full range of literary understanding, including cultural interpretation, open-ended metaphor analysis, and aesthetic judgment. Finally, because modern Chinese poetry differs substantially from classical Chinese poetry and poetry in other languages and traditions, our findings should not be generalized to genres such as classical Chinese poetry, haiku, or Arabic poetry without further study.

\section*{Ethics Statement}

This study collects poetic samples exclusively from contemporary poets, and all data are used strictly for non-commercial scientific research purposes. We hope that this approach will foster the integration of AI and literature.

Furthermore, all experts involved were fairly compensated at rates above the minimum wage, ensuring that our research was conducted in full compliance with legal standards.


\bibliography{acl_latex}
\newpage
\appendix
\begin{CJK}{UTF8}{gbsn}
\label{sec:appendix}

\section{Annotation Quality Control}
\label{app:annotation-quality}

DeepSeek-V4-Pro was used only to assist mask selection and initial distractor generation for IMF and SNI. Professional poets reviewed every candidate and substantially revised or replaced the vast majority of generated distractors before inclusion. For STO and LIO, experts removed options with obvious discontinuities or contradictory imagery and redesigned them so that the contrast depended on internal structure and progression. For IMF, they constrained candidates to concrete imagery nouns and eliminated cases with more than one plausible answer. For SNI, they ensured that candidates were complete sentences of similar length and rewrote distractors that merely repeated the same imagery with superficial changes. All final answers and options were determined by the experts.

\section{Prompts We Used in Peony}

\label{app:prompts}
\subsection{Prompts used in Data Curation}
During the data curation process, we employed DeepSeek-V4-Pro as an annotation assistant. Its primary task was to select appropriate imagery and sentences for blanking in the IMF and SNI tasks, as well as to generate distractor options. It should be emphasized that this step was intended solely to reduce the manual effort of human experts. All final items were completed under the review and revision of human experts. The prompts used for DeepSeek-V4-Pro can be found in Table \ref{tab:prompt-imf-curation} and \ref{tab:prompt-sni-curation}. 
Where the \textit{Candidate Imagery Word Pool} (\texttt{\{noun\_pool\}}) is chosen from the nouns existing in the given poem, and \texttt{\{num\_to\_dig\}} depends on the length of the poem. 
\begin{table*}[ht]
\centering
\small
\begin{tcolorbox}[colback=gray!5, colframe=black, width=\textwidth]
\textbf{Prompt for Curating the Imagery Filling Task} \\

你是一个中国现代诗歌研究专家。请阅读这首诗《\{title\}》：
\{content\}

【可选意象实词库】：
\{noun\_pool\}

任务：
从【可选意象实词库】中选出 \texttt{\{num\_to\_dig\}} 个核心意象词进行挖空并出题。

严格执行如下说明：

\begin{enumerate}[noitemsep, topsep=0pt]
    \item 必须是“看得见、摸得着的实体”：
    \begin{enumerate}[noitemsep, topsep=0pt]
        \item 允许：[陶罐]、[铠甲]、[容器]、[落叶]、[飞鸟]、[铁轨]、[窗棂]。
        \item 禁止（抽象名词/状态）：[羁绊]、[帷幕]、[深渊]、[白色]、[表情]、[符号]、[年轮]、[守望]、[背影]。
        \item 禁止（生理/空间概念）：[褶皱]、[沟壑]、[触感]、[距离]。
    \end{enumerate}
    
    \item 必须是“单纯的静止物”：
    \begin{enumerate}[noitemsep, topsep=0pt]
        \item 禁止（动名组合/动作化名词）：[燃香]、[落日]、[余晖]、[呼啸]、[呐喊]。
        \item 禁止（单纯的量词）：[一瓣]、[一片]、[一缕]。
    \end{enumerate}
    
    \item 宁缺毋滥原则：
    \begin{enumerate}[noitemsep, topsep=0pt]
        \item 如果某个名词你不确定它是否属于“物理实体”，请坚决放弃。
        \item 只选择你 100\% 确定是具体物品、实际事物的名词。
    \end{enumerate}
    
    \item 干扰项生成逻辑：
    \begin{enumerate}[noitemsep, topsep=0pt]
        \item 必须是同样属性的“物理实体名词”。
        \item 如果原词是[陶罐]，干扰项应为[瓷瓶]、[木桶]、[瓦罐]等实体物，严禁出现抽象词。
    \end{enumerate}
\end{enumerate}

请严格按格式输出：

[QUESTION]
（挖空后的文本，使用【空1】、【空2】标识）

[ANSWERS]
（原词1）|（原词2）

[OPTIONS]
【空1】:（干扰1）|（干扰2）|（干扰3）
【空2】:（干扰1）|（干扰2）|（干扰3）

\textbf{English Translation}\\
You are an expert in modern Chinese poetry. Please read the following poem titled "\{title\}":

\{content\}

[Candidate Imagery Word Pool]:
\{noun\_pool\}

Task:
Select \texttt{\{num\_to\_dig\}} core imagery words from the candidate pool to create cloze test style questions.

Strict Rules for Imagery Substantiation:

\begin{enumerate}[noitemsep, topsep=0pt]
    \item Must be "visible and tangible entities":
    \begin{enumerate}[noitemsep, topsep=0pt]
        \item Allowed: [earthen jar], [armor], [container], [fallen leaf], [bird], [railway track], [window lattice].
        \item Forbidden (abstract nouns/states): [bond], [curtain], [abyss], [whiteness], [expression], [symbol], [tree ring], [vigil], [back view].
        \item Forbidden (physiological/spatial concepts): [folds], [grooves], [touch], [distance].
    \end{enumerate}
    
    \item Must be "static objects":
    \begin{enumerate}[noitemsep, topsep=0pt]
        \item Forbidden (verb-noun compounds/action nouns): [burning incense], [setting sun], [afterglow], [howling], [shout].
        \item Forbidden (pure measure words): [a petal], [a slice], [a wisp].
    \end{enumerate}
    
    \item Better to omit than to force:
    \begin{enumerate}[noitemsep, topsep=0pt]
        \item If you are uncertain whether a noun qualifies as a "physical entity," discard it.
        \item Only select nouns you are 100\% certain are concrete objects.
    \end{enumerate}
    
    \item Distractor Generation Logic:
    \begin{enumerate}[noitemsep, topsep=0pt]
        \item Distractors must also be "physical entity nouns" of the same category.
        \item For example, if the correct answer is [earthen jar], distractors should be [porcelain vase], [wooden bucket], [clay pot] — abstract words are strictly prohibited.
    \end{enumerate}
\end{enumerate}

Output strictly in the following format:

[QUESTION]
(Text with blanks, marked as 【Blank1】, 【Blank2】)

[ANSWERS]
(word1)|(word2)

[OPTIONS]
【Blank1】:(distractor1)|(distractor2)|(distractor3)
【Blank2】:(distractor1)|(distractor2)|(distractor3)

\end{tcolorbox}
\caption{Prompt for Curating the Imagery Filling Task}
\label{tab:prompt-imf-curation}
\end{table*}

\begin{table*}[ht]
\centering
\small
\begin{tcolorbox}[colback=gray!5, colframe=black, width=\textwidth]
\textbf{Prompt for Curating the Sentence Inference Task} \\

你是一个中国现代诗歌研究专家。请阅读以下诗歌《\{title\}》：

\{content\}

任务：
从这首诗中选出 \texttt{\{num\_to\_dig\}} 个关键句子进行挖空，并为每个空位生成3个合理的干扰选项。

严格执行【句子选择与干扰项生成规范】：

\begin{enumerate}[noitemsep, topsep=0pt]
    \item 关键句子选择标准：
    \begin{enumerate}[noitemsep, topsep=0pt]
        \item 优先选择对诗歌意境推进、情感转折或逻辑连贯性起核心作用的句子。
        \item 所选句子应具备一定的语义独立性和可替换性。
    \end{enumerate}
    
    \item 干扰项生成逻辑：
    \begin{enumerate}[noitemsep, topsep=0pt]
        \item 干扰句应与正确答案在长度和句式结构上相近。
        \item 干扰句与诗歌主题或语境有一定关联，但填入后会破坏情感走向、叙事逻辑或意境连贯性。
        \item 干扰句应具有一定的迷惑性，不能明显与诗歌无关。
    \end{enumerate}
    
    \item 挖空位置标记：
    \begin{enumerate}[noitemsep, topsep=0pt]
        \item 使用【空1】、【空2】、【空3】等标识替换被选中的句子。
        \item 保持诗歌原文的其他部分不变。
    \end{enumerate}
    
    \item 输出格式要求：
    \begin{enumerate}[noitemsep, topsep=0pt]
        \item 严格按照下方格式输出，不要输出其他内容。
    \end{enumerate}
\end{enumerate}

请严格按格式输出：

[QUESTION]
（带有【空X】标识的诗歌全文）

[ANSWERS]
（原句1）|（原句2）|（原句3）

[OPTIONS]
【空1】:（干扰句A）|（干扰句B）|（干扰句C）
【空2】:（干扰句A）|（干扰句B）|（干扰句C）
【空3】:（干扰句A）|（干扰句B）|（干扰句C）

\textbf{English Translation}\\
You are an expert in modern Chinese poetry. Please read the following poem titled "\{title\}":

\{content\}

Task:
Select \texttt{\{num\_to\_dig\}} key sentences from this poem to create blanks, and generate three reasonable distractors for each blank.

Strict Rules for Sentence Selection and Distractor Generation:

\begin{enumerate}[noitemsep, topsep=0pt]
    \item Key Sentence Selection Criteria:
    \begin{enumerate}[noitemsep, topsep=0pt]
        \item Prioritize sentences that play a central role in advancing the poetic mood, marking emotional transitions, or maintaining logical coherence.
        \item Selected sentences should have a certain degree of semantic independence and substitutability.
    \end{enumerate}
    
    \item Distractor Generation Logic:
    \begin{enumerate}[noitemsep, topsep=0pt]
        \item Distractors should be similar in length and syntactic structure to the correct answer.
        \item Distractors should be semantically relevant to the poem's theme or context, but would disrupt emotional progression, narrative logic, or mood coherence if inserted.
        \item Distractors should be plausible and challenging, not obviously irrelevant to the poem.
    \end{enumerate}
    
    \item Blank Placement:
    \begin{enumerate}[noitemsep, topsep=0pt]
        \item Replace each selected sentence with 【Blank1】, 【Blank2】, 【Blank3】, etc.
        \item Keep the rest of the poem unchanged.
    \end{enumerate}
    
    \item Output Format:
    \begin{enumerate}[noitemsep, topsep=0pt]
        \item Strictly follow the format below without any additional content.
    \end{enumerate}
\end{enumerate}

Output strictly in the following format:

[QUESTION]
(Full poem text with 【BlankX】 markers)

[ANSWERS]
(original sentence1)|(original sentence2)|(original sentence3)

[OPTIONS]
【Blank1】:(distractor A)|(distractor B)|(distractor C)
【Blank2】:(distractor A)|(distractor B)|(distractor C)
【Blank3】:(distractor A)|(distractor B)|(distractor C)

\end{tcolorbox}
\caption{Prompt for Curating the Sentence Inference Task}
\label{tab:prompt-sni-curation}
\end{table*}

\subsection{Prompts used in Evaluation}
\begin{table*}[ht]
\centering
\begin{tcolorbox}[colback=gray!5, colframe=black, width=\textwidth]
\textbf{Prompt for Stanza Ordering Task ($P_{STO}$):} \\
请阅读以下被打乱顺序的诗歌片段，完成诗节排序恢复任务。每个带有编号的片段代表诗歌中一个相对完整的独立诗节。你需要通过分析诗节之间的叙事推进、情绪递进和意象演变，从候选选项中选出最合理的完整诗歌结构顺序。

【待排序的完整诗歌】：
\{question\_text\}

【候选选项（代表诗节编号的正确前后顺序）】：
\{options\_text\}

【任务要求】：
请先简要分析诗节间的逻辑关联，最后，必须严格按照以下格式在一行内输出最终答案，不要有多余字符：
答案是：X （X代表你选择的选项对应的大写字母）

\vspace{0.8em}
\textbf{English Translation}\\
Please read the following scrambled poetry excerpts and complete the stanza ordering restoration task. Each numbered excerpt represents a relatively complete independent stanza. Analyze the narrative progression, emotional development, and imagery evolution between stanzas to select the most reasonable complete poetic structure from the candidate options.

[Poetry Content to be Ordered]:
\{question\_text\}

[Candidate Options (representing the correct order of stanza numbers)]:
\{options\_text\}

[Task Requirements]:
Provide a brief analysis of the logical connections between stanzas. Finally, output the final answer in a single line strictly following the format below without any extra characters:
Answer: X (X represents a single uppercase letter corresponding to the option you have selected)
\end{tcolorbox}
\caption{Prompt for Stanza Ordering (STO) Task}
\label{tab:prompt-sto}
\end{table*}

\begin{table*}[ht]
\centering
\begin{tcolorbox}[colback=gray!5, colframe=black, width=\textwidth]
\textbf{Prompt for Sentence Ordering Task ($P_{LIO}$):} \\
请阅读以下诗歌片段，包含一部分正确顺序的诗歌和待排序的片段，完成诗句排序任务，从候选选项中选出最合理的完整诗句顺序。

【待排序的完整诗歌】：
\{question\_text\}

【候选选项（代表诗句编号的正确前后顺序）】：
\{options\_text\}

【任务要求】：
请先简要分析，最后，必须严格按照以下格式在一行内输出最终答案，不要有多余字符：
答案是：X （X代表你选择的单个大写字母）

\vspace{0.8em}
\textbf{English Translation}\\
Please read the following poetry excerpt, which includes a portion of correctly ordered lines and fragments to be sorted. Complete the sentence ordering task by selecting the most reasonable complete sequence from the candidate options.

[Poetry Content to be Ordered]:
\{question\_text\}

[Candidate Options (representing the correct order of line numbers)]:
\{options\_text\}

[Task Requirements]:
Provide a brief analysis first. Finally, output the final answer in a single line strictly following the format below without any extra characters:
Answer: X (X represents a single uppercase letter corresponding to the option you have selected)
\end{tcolorbox}
\caption{Prompt for Sentence Ordering (LIO) Task}
\label{tab:prompt-lio}
\end{table*}

\begin{table*}[ht]
\centering
\begin{tcolorbox}[colback=gray!5, colframe=black, width=\textwidth]
\textbf{Prompt for Imagery Filling Task ($P_{IMF}$):} \\
请阅读以下诗歌，并完成诗歌完形填空选择题。你需要通过分析诗歌的意境、叙事逻辑和情感走向，从候选选项中选出最符合原诗空缺处的正确答案。

【诗歌标题】：
\{poem\_title\}

【带有空缺处的完整诗歌】：
\{question\_text\}

【候选选项列表】：
\{options\_text\}

【任务要求】：
请先简要分析空缺处的逻辑和意境，最后，必须严格按照以下格式在一行内输出最终答案，不要有多余字符：
如果有1个空：答案是：X
如果有2个空（按空1、空2顺序连写）：答案是：XY

\vspace{0.8em}
\textbf{English Translation}\\
Please read the following poem and complete the cloze test multiple-choice questions. Analyze the poetic conception, narrative logic, and emotional direction to select the correct answer that best fits each blank.

[Poem Title]:
\{poem\_title\}

[Full Poem with Blanks]:
\{question\_text\}

[Candidate Options List]:
\{options\_text\}

[Task Requirements]:
Provide a brief analysis of the logic and poetic conception at each blank. Finally, output the final answer in a single line strictly following the format below without any extra characters:
If there is one blank: Answer: X
If there are two blanks (concatenated in order of blank 1, blank 2): Answer: XY
\end{tcolorbox}
\caption{Prompt for Imagery Filling (IMF) Task}
\label{tab:prompt-imf}
\end{table*}

\begin{table*}[ht]
\centering
\begin{tcolorbox}[colback=gray!5, colframe=black, width=\textwidth]
\textbf{Prompt for Sentence Inference Task ($P_{SNI}$):} \\
请阅读以下诗歌，完成句子推断任务。你需要从每个空位的候选项中选出最符合该处语境的句子。

【完整诗歌】：
\{question\_text\}

【候选选项】：
\{options\_text\}

【任务要求】：
请先简单写出你的分析过程。
最后，必须严格按照以下格式在一行内输出最终答案，不要有多余字符：
答案是：XYZ （有几个空就写几个字母）

\vspace{0.8em}
\textbf{English Translation}\\
Please read the following poem and complete the sentence inference task. Select the sentence that best fits the context for each blank from the candidate options.

[Full Poem]:
\{question\_text\}

[Candidate Options]:
\{options\_text\}

[Task Requirements]:
Briefly write your analysis process.
Finally, output the final answer in a single line strictly following the format below without any extra characters:
Answer: XYZ (Write as many letters as there are blanks)
\end{tcolorbox}
\caption{Prompt for Sentence Inference (SNI) Task}
\label{tab:prompt-sni}
\end{table*}
For the four evaluation tasks in Peony, we designed task-specific prompts to ensure consistent and reproducible model assessment. Each prompt includes a clear task description, the input format (full Poem or excerpt with blanks), candidate options, and strict output formatting requirements to facilitate automatic evaluation. The complete prompts can be found in Table \ref{tab:prompt-sto}, \ref{tab:prompt-lio}, \ref{tab:prompt-imf} and \ref{tab:prompt-sni}.

\section{Supplementary Analyses and Benchmark Comparison}
\label{app:supplementary-analysis}

Table~\ref{tab:answer-distribution} shows that correct answers are approximately balanced across positions A--D in every task.

\begin{table}[H]
\centering
\small
\rowcolors{2}{white}{tablerowcolor}
\resizebox{\columnwidth}{!}{%
\begin{tabular}{lrrrrr}
\toprule
\rowcolor{tableheadcolor}
\textbf{Task} & \textbf{Answers} & \textbf{A (\%)} & \textbf{B (\%)} & \textbf{C (\%)} & \textbf{D (\%)} \\
\midrule
IMF & 695 & 27.5 & 25.0 & 23.5 & 24.0 \\
SNI & 2,424 & 25.4 & 24.1 & 25.7 & 24.7 \\
LIO & 800 & 23.6 & 25.9 & 26.0 & 24.5 \\
STO & 436 & 24.5 & 24.5 & 25.0 & 25.9 \\
Overall & 4,355 & 25.3 & 24.6 & 25.4 & 24.7 \\
\bottomrule
\end{tabular}}
\caption{Distribution of correct answer positions.}
\label{tab:answer-distribution}
\end{table}

To examine positional bias further, we randomly permuted each STO question into two additional option orders (STO1 and STO2) and re-evaluated two representative non-thinking models. As shown in Table~\ref{tab:position-bias}, performance remains stable across option orders.

\begin{table}[H]
\centering
\small
\resizebox{\columnwidth}{!}{%
\begin{tabular}{lrrr}
\hline
\rowcolor{blue!15} \textbf{Model} & \textbf{Original} & \textbf{STO1} & \textbf{STO2} \\
\hline
DeepSeek-V4-Flash & 47.02 & 47.71 & 46.56 \\
\rowcolor{blue!5} GPT-5.1 & 39.08 & 39.45 & 39.68 \\
\hline
\end{tabular}}
\caption{STO accuracy under two random option permutations.}
\label{tab:position-bias}
\end{table}

Most existing Chinese poetry benchmarks focus on classical poetry. Table~\ref{tab:benchmark-comparison} situates Peony as a modern-poetry benchmark targeting poetic-logic comprehension rather than generation detection or a single semantic classification task.

\begin{table*}[t]
\centering
\small
\rowcolors{2}{white}{tablerowcolor}
\resizebox{\textwidth}{!}{%
\begin{tabular}{llcll}
\toprule
\rowcolor{tableheadcolor}
\textbf{Benchmark} & \textbf{Poetry Type} & \textbf{\#Tasks} & \textbf{Dataset Scale} & \textbf{Evaluation Target} \\
\midrule
CCPM (Li et al., 2021) & Classical & 1 & 27,218 poems & Semantic matching \\
Hou et al. (2025) & Classical & 1 & 2,918 poems & Theme classification \\
ChangAn (Li et al., 2026) & Classical & 4 & 10K+ human-written + 20K+ AI-generated & LLM-generated poetry detection \\
Neo-Classic (Zhang et al., 2026) & Classical & 5 & 1,406 poems (628 Shi + 778 Ci) & Linguistic-aesthetic reasoning \\
AIGenPoetry (Wang et al., 2025) & Modern & 1 & 800 human-written + 41,600 LLM-generated & LLM-generated poetry detection \\
Peony & Modern & 4 & 800 poems / 4,355 instances & Poetic-logic comprehension \\
\bottomrule
\end{tabular}}
\caption{Comparison with representative Chinese poetry benchmarks.}
\label{tab:benchmark-comparison}
\end{table*}

\section{Imagery in Peony}
\label{app:imagery}

Imagery constitutes a cornerstone of modern Chinese poetry, serving as the primary vehicle through which poets convey emotion, evoke atmosphere, and construct poetic meaning. Poetic imagery in modern Chinese poetry also carries rich connotations, relies on unconventional associations, and participates in complex networks of semantic resonance across lines and stanzas. 

To characterize the imagery landscape of Peony, we perform a statistical analysis of imagery word frequencies across the 800 poems in our benchmark. We distinguish between short imagery words (single-character words) and long imagery words (two-character words). Table \ref{tab:human-imagery-freq} presents the top 20 high-frequency imagery words in Peony, divided into short and long forms.

As shown in Table \ref{tab:human-imagery-freq}, the most frequent short imagery word is "人 (Human)" with 245 occurrences, followed by "爱 (Love)" (90) and "风 (Wind)" (68). Natural elements such as "水 (Water)", "雨 (Rain)", and "雪 (Snow)" also appear prominently, reflecting the enduring influence of classical nature-themed poetry on contemporary practice. For long imagery words, "自己 (Self)" ranks highest (145), followed by "天空 (Sky)" (99) and "生命 (Life)" (89). Temporal concepts including "岁月 (Time)", "时间 (Time)", and "季节 (Season)" also show high frequency, indicating that the passage of time and seasonal change serve as recurring thematic concerns.

Notably, the presence of abstract yet tangible terms such as "人间 (Human World)" and "世界 (World)" suggests that modern Chinese poetry engages not only with concrete sensory experiences but also with philosophical reflections on existence and society. The diversity and richness of these imagery words collectively establish Peony as a challenging benchmark for evaluating LLMs' capacity to comprehend poetic logic.

\begin{table*}[t]
\centering
\rowcolors{2}{white}{tablerowcolor}
\resizebox{0.8\textwidth}{!}{
\setlength{\tabcolsep}{10pt}
\begin{tabular}{rlc | rlc}
\toprule
\rowcolor{tableheadcolor}
\multicolumn{3}{c|}{\textbf{Top 20 Short Imagery Words}} & \multicolumn{3}{c}{\textbf{Top 20 Long Imagery Words}} \\
\cmidrule(lr){1-3} \cmidrule(lr){4-6}
\rowcolor{tableheadcolor}
\textbf{Rank} & \textbf{Imagery} & \textbf{Count} & \textbf{Rank} & \textbf{Imagery} & \textbf{Count} \\ 
\midrule
1 & 人 (Human) & 245 & 1 & 自己 (Self) & 145 \\
2 & 爱 (Love) & 90 & 2 & 天空 (Sky) & 99 \\
3 & 风 (Wind) & 68 & 3 & 生命 (Life) & 89 \\
4 & 梦 (Dream) & 46 & 4 & 雨水 (Rainwater) & 80 \\
5 & 水 (Water) & 45 & 5 & 阳光 (Sunlight) & 73 \\
6 & 雨 (Rain) & 44 & 6 & 城市 (City) & 68 \\
7 & 诗 (Poem) & 41 & 7 & 石头 (Stone) & 65 \\
8 & 花 (Flower) & 34 & 8 & 月光 (Moonlight) & 64 \\
9 & 城 (City) & 32 & 9 & 人间 (Human World) & 63 \\
10 & 声 (Sound) & 30 & 10 & 世界 (World) & 62 \\
11 & 雪 (Snow) & 28 & 11 & 岁月 (Time) & 61 \\
12 & 写 (Write) & 26 & 12 & 流水 (Flowing) & 59 \\
13 & 年 (Year) & 26 & 13 & 身体 (Body) & 53 \\
14 & 云 (Cloud) & 25 & 14 & 文字 (Words) & 52 \\
15 & 夜 (Night) & 25 & 15 & 时间 (Time) & 51 \\
16 & 山 (Mountain) & 24 & 16 & 大地 (Earth) & 50 \\
17 & 月 (Moon) & 24 & 17 & 春天 (Spring) & 48 \\
18 & 新 (New) & 24 & 18 & 季节 (Season) & 48 \\
19 & 光 (Light) & 23 & 19 & 落日 (Sunset) & 46 \\
20 & 树 (Tree) & 21 & 20 & 声音 (Sound) & 45 \\
\bottomrule
\end{tabular}}
\caption{Top 20 high-frequency short and long imagery words in Peony.}
\label{tab:human-imagery-freq}
\end{table*}


\section{Examples of Different Tasks}
\label{app:examples-of-poetry}
Table \ref{Experimental examples} provides concrete examples of the Stanza Ordering (STO), Line Ordering (LIO), Imagery Filling (IMF), and Sentence Inference (SNI) tasks. Only one example is shown per task.

\begin{table*}[ht]
\small
\begin{tcolorbox}[colback=gray!5,
                  colframe=black,
                  width=0.95\textwidth]

\textbf{Stanza Ordering (STO)}\\
    \textbf{title:} 《庄子与倒影》
    
    \textbf{Original Content:} 梦见蝴蝶的人，也梦见了巨鸟和长鲸/它们一样轻盈/一样翩跹，在他身体里//身着羽衣、袈裟或者儒冠/一个人，竟可以是任何事物/而我，终于用阅读//追上了自己的前身。漆园吏，漆园吏/生活的迷宫越来越窄，仙术废弛/夜晚的故事不再发生//漆园吏，漆园吏，红尘里/尽是马不停蹄的人，每一个/都是你秋水中的倒影
    
    \textbf{Question:} 1. 漆园吏，漆园吏，红尘里/尽是马不停蹄的人，每一个/都是你秋水中的倒影//2. 梦见蝴蝶的人，也梦见了巨鸟和长鲸/它们一样轻盈/一样翩跹，在他身体里//3. 追上了自己的前身。漆园吏，漆园吏/生活的迷宫越来越窄，仙术废弛/夜晚的故事不再发生//4. 身着羽衣、袈裟或者儒冠/一个人，竟可以是任何事物/而我，终于用阅读
    
    \textbf{Options:} A.2431  B.1432  C.4231  D.2341
    
    \textbf{Correct Answer:} A
\vspace{0.6em}\\

\hrule
\vspace{0.6em}
\textbf{Line Ordering (LIO)} \\
\textbf{title: 《伤痕》} \\
\textbf{Original Content:} 左手虎口处，有一个刀疤/类似外公的荒坟，落魄而寒酸/又像一个残坑/不知曾经栽种过什么//那是童年跟着外公干活/刻下的记忆，一块小伤疤/凹成一个温暖的小窝/时常召唤出走多年的故人/回归

\textbf{Question:} 左手虎口处，有一个刀疤/类似外公的荒坟，落魄而寒酸/又像一个残坑/不知曾经栽种过什么//1. 那是童年跟着外公干活/2. 时常召唤出走多年的故人/3. 回归/4. 凹成一个温暖的小窝/5. 刻下的记忆，一块小伤疤

\textbf{Options:} A.13425    B.15423    C.53142    D.25413

\textbf{Correct Answer:} B
\vspace{0.6em}\\

\hrule
\vspace{0.6em}

\textbf{Imagery Filling (IMF)}

    \textbf{title:} 《晚秋帖》
    
    \textbf{Original Content:} 秋风在远处放着红叶的血。晚秋/执行着天地的理性，一身肃杀之气/回眸就让月光结霜，也让白草生露/为九月菊镀上金身，也误尽秋虫的一生/这是祈祷的时刻，万物在等候发落/有人落叶纷飞，有人在梧桐树下/不安地游荡。只有你献出了白雪/眼中响着春天的轻雷。我看见/我们的爱情端坐于白云的寓体之上/天宇澄清，河水澄清
    
    \textbf{Question:} 秋风在远处放着【空1】的血。晚秋/执行着天地的理性，一身肃杀之气/回眸就让【空2】结霜，也让白草生露/为九月菊镀上金身，也误尽【空3】的一生/这是祈祷的时刻，万物在等候发落/有人落叶纷飞，有人在梧桐树下/不安地游荡。只有你献出了白雪/眼中响着春天的轻雷。我看见/我们的爱情端坐于白云的寓体之上/天宇澄清，河水澄清
    
    \textbf{Options:}  
    
      【空1】: 
        A.桑叶  B.银杏  C.晚霞  D.红叶

      【空2】: 
        A.月光  B.麦秆  C.土地  D.乌鸦
        
      【空3】: 
        A.蚯蚓
        B.秋虫
        C.蜻蜓
        D.蚂蚁

    \textbf{Correct Answer:} DAB
\vspace{0.6em}\\

\hrule
\vspace{0.6em}

\textbf{Sentence Inference (SNI)} 

    \textbf{title:} 《一棵草》
    
    \textbf{Question:} 一棵草，站在白色的坟头/瘦如墓碑/它被秋天熄灭/【空1】/季节有时候大过马蹄/有时候浅过蟋蟀的眼睛/【空2】/【空3】/它也会祭奠自己/仿佛久不归家之人/在月圆之夜/顿首遥拜
    
    \textbf{Options:}  
    
      【空1】: 
        A.却被霜雪掩埋
        B.又被夏日遗忘
        C.又在风中苏醒
        D.又被春天点燃

      【空2】: 
        A.谷雨前后，如果
        B.端午之前，如果
        C.清明的时候，如果
        D.立春之后，如果
        
      【空3】: 
        A.雨还未至
        B.荷花未开
        C.无人到来
        D.积雪未消
      
    \textbf{Correct Answer:} DCC\\

\end{tcolorbox}
\caption{Experimental examples for different tasks. / represents a line break, // represents a stanza break.}
\label{Experimental examples}
\end{table*}

\section{Human Performance}
\label{app:human-performance}

We recruited five professional poets and four non-expert readers to provide a human reference on a random sample of 40 poems. None of the professional poets authored poems in Peony. The non-experts had received Chinese-language education at the undergraduate level but had not formally studied modern Chinese poetry. The experts had more than eight years of poetry-writing experience, had published individual poetry collections, and had published a substantial number of modern Chinese poems. Table~\ref{tab:human-performance} reports Accuracy for all four tasks and \textit{PR} in parentheses for IMF and SNI. Experts achieve a mean score of 51.44, compared with 41.28 for non-experts, with substantial variation within both groups.

\begin{table*}[t]
\centering
\scriptsize
\resizebox{0.85\textwidth}{!}{%
\begin{tabular}{lccccc}
\hline
\rowcolor{blue!15} \textbf{Reader} & \textbf{IMF (\textit{PR})} & \textbf{SNI (\textit{PR})} & \textbf{LIO} & \textbf{STO} & \textbf{Avg.} \\
\hline
Expert 1 & 94.12 (90) & 70 (40) & 40 & 80 & 71.03 \\
\rowcolor{blue!5} Expert 2 & 47.06 (20) & 16.67 (0) & 60 & 60 & 45.93 \\
Expert 3 & 82.35 (70) & 30 (10) & 50 & 30 & 48.09 \\
\rowcolor{blue!5} Expert 4 & 58.82 (40) & 36.67 (10) & 50 & 20 & 41.37 \\
Expert 5 & 76.47 (60) & 36.67 (10) & 80 & 10 & 50.79 \\
\hline
\rowcolor{blue!5} Non-expert 1 & 70.59 (50) & 30 (10) & 60 & 40 & 50.15 \\
Non-expert 2 & 70.59 (60) & 26.67 (0) & 40 & 30 & 41.82 \\
\rowcolor{blue!5} Non-expert 3 & 47.06 (10) & 26.67 (0) & 30 & 50 & 38.43 \\
Non-expert 4 & 58.82 (40) & 20 (0) & 30 & 30 & 34.71 \\
\hline
\rowcolor{blue!5}
All readers & 67.32 & 32.59 & 48.89 & 38.89 & 46.92 \\
Expert mean & 71.76 & 38 & 56 & 40 & 51.44 \\
\rowcolor{blue!5} Non-expert mean & 61.77 & 25.84 & 40 & 37.5 & 41.28 \\
\hline
\end{tabular}}
\caption{Human performance. \textit{PR} is shown in parentheses.}
\label{tab:human-performance}
\end{table*}

\section{SNI Case Study}
\label{app:sni-case-study}

SNI is the most challenging task. For the poem ``Lintie Ji'' (Copying Calligraphy), DeepSeek-V4-Flash in thinking mode answers both blanks incorrectly (AA rather than DC). The first error shows a preference for locally compatible ancient imagery: the model chooses the bamboo-slip and ox-cart option, but overlooks the action-level continuity between ``leaping onto horseback'' in the correct option and ``passing through a rain of arrows'' in the following line. The second error shows that it does not connect ``one dot, one stroke, one hook'' with calligraphic strokes and therefore misses the option containing ``xuan paper.'' The reasoning also recognizes the allusion to Wang Xizhi's \textit{Preface to the Orchid Pavilion Collection} but fails to use its background correctly, treating ``Orchid Pavilion winding stream'' as the relevant location rather than identifying ``Kuaiji Shanyin.'' This example illustrates how surface imagery matching, cross-line logic, and incomplete use of cultural knowledge can jointly produce SNI errors.

\begin{table*}[ht]
\centering
\small
\begin{tcolorbox}[colback=gray!5,colframe=black,width=0.97\textwidth]
\textbf{Title:} 《临帖记》 (Copying Calligraphy)\\
\textbf{Question:} 天用云作字。地，沉埋龙骨与钟鼎/【Blank 1】/穿过箭雨，找寻真书的筋骨/山谷间长蛇吐着信子，东坡/有丑石压住蛤蟆。仙鹤掠过琉璃瓦/故国如梦，笔给自己的冢/题写碑文。古人隐进了时间的群山/【Blank 2】/一点，一撇，一弯钩，脚印合着脚印/墨是谁家苍苔，熬过了结冰季节/于深春时爬上我的袖口\\[0.4em]
\textbf{Blank 1 options:}\\
A. 流水运笔至兰亭曲水，仍有竹简般的纹理。汉字跃上牛车\\
B. 流水运笔至长安城下，仍有碑林般的肃穆。汉字跃上驼峰\\
C. 流水运笔至浔阳江头，仍有琵琶般的弦音。汉字跃上船头\\
D. 流水运笔至会稽山阴，仍有丝绸般的波纹。汉字跃上马背\\[0.4em]
\textbf{Blank 2 options:}\\
A. 只在砚池的深潭里藏起鳞光\\
B. 只在青石的小径上留下屐痕\\
C. 只在宣纸的雪地里留下足迹\\
D. 只在绢帛的秋水中隐去面容\\[0.4em]
\textbf{Correct answers:} DC \qquad \textbf{DeepSeek-V4-Flash (Thinking):} AA
\end{tcolorbox}
\caption{An SNI case study illustrating cross-line poetic-logic errors.}
\label{tab:sni-case-study}
\end{table*}

\end{CJK}

\end{document}